\documentclass[10pt,twocolumn,letterpaper]{article}

\usepackage[]{cvpr}      

\usepackage{tcolorbox}
\usepackage{multirow}
\usepackage[table]{xcolor}
\usepackage{tabularx, colortbl}
\usepackage{caption}

\newcolumntype{Y}{>{\raggedright\arraybackslash}X}
\definecolor{shadecolor}{gray}{0.9}

\usepackage{soul}
\definecolor{RedOrange}{RGB}{255,69,0}
\definecolor{LGray2}{gray}{0.92}
\sethlcolor{LGray2}

\newcommand{\insight}[1]{\hl{#1}}
\usepackage{multirow}

\definecolor{cvprblue}{rgb}{0.21,0.49,0.74}
\usepackage[pagebackref,breaklinks,colorlinks,allcolors=cvprblue]{hyperref}

\def\paperID{5} 
\def\confName{CVPR}
\def\confYear{2026}

\title{BareBones: Benchmarking Zero-Shot Geometric Comprehension in VLMs}

\author{
Aaditya Baranwal*\\
University of Central Florida\\
{\tt\small aaditya.baranwal@ucf.edu}
\and
Vishal Yadav*\\
(Independent Researcher)\\
{\tt\small vishalvy789@gmail.com}
\and
Abhishek Rajora\\
University of Calgary\\
{\tt\small abhishek.rajora@ucalgary.edu.ca}
}

\makeatletter
\AtBeginDocument{
\let\@oldmaketitle\@maketitle
\renewcommand{\@maketitle}{\@oldmaketitle
  \centering
  \centerline{\includegraphics[width=0.9\linewidth]{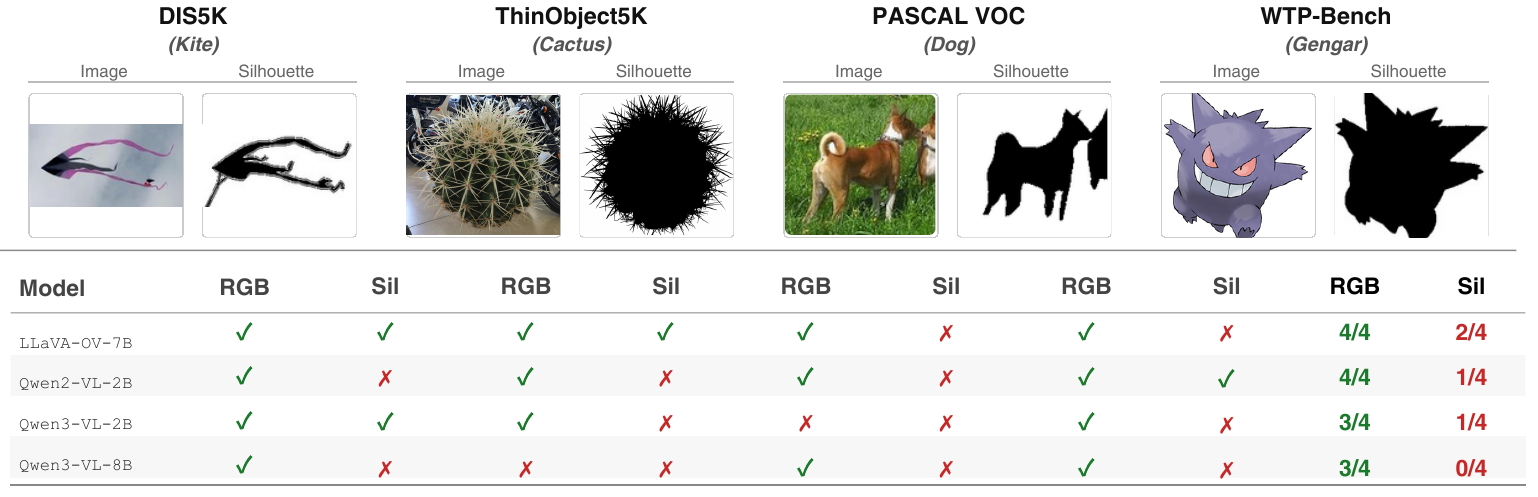}}
  \captionof{figure}{
    \textbf{RGB vs.\ Silhouette-only Recognition Across Benchmark Subsets.} Representative image--silhouette pairs from four datasets evaluated against four VLMs under two conditions: standard RGB input and silhouette-only inputs. Check marks (\textcolor{green}{\checkmark}) and crosses (\textcolor{red}{$\times$}) denote correct and incorrect top-1 predictions, respectively. Aggregate scores (rightmost columns) reveal a consistent and severe drop in identification accuracy under complete RGB deprivation, corroborating the \textit{Texture Bias Cliff} hypothesis.
  \label{fig:teaser}
 }
 \vspace{2mm}
}
}
\makeatother

\begin{document}
\maketitle
\begin{abstract}
While Vision-Language Models (VLMs) demonstrate remarkable zero-shot recognition capabilities across a diverse spectrum of multimodal tasks, it yet remains an open question whether these architectures genuinely comprehend geometric structure or merely exploit RGB textures and contextual priors as statistical shortcuts. Existing evaluations fail to isolate this mechanism, conflating semantic reasoning with texture mapping and relying on imprecise annotations that inadvertently leak environmental cues. To address this gap, we introduce \textbf{BareBones}, a zero-shot benchmark designed to stress-test pure geometric shape comprehension. We curate pixel-level silhouettes of geometrically distinct classes across six datasets: five established segmentation sources (ImageNet-S, DIS5K, ThinObject5K, PASCAL VOC, CUB-200) and our novel flagship collection, WTP-Bench, establishing a noise-free geometric taxonomy. WTP-Bench is an extreme, fine-grained visual puzzle that forces models to identify inter-class geometric concepts from boundary contours alone. Our evaluation of 26 state-of-the-art proprietary and open-weight VLMs (\eg, GPT-4.1, Gemini, Claude Sonnet 4.5, LLaVA) reveals a consistent, severe performance collapse under RGB deprivation, a phenomenon we term the \textit{Texture Bias Cliff}. By documenting universal structural blindspots, BareBones establishes a rigorous yardstick for genuine geometric grounding.
\end{abstract}

\begin{figure*}[!t]
    \centering
    \includegraphics[width=0.85\linewidth]{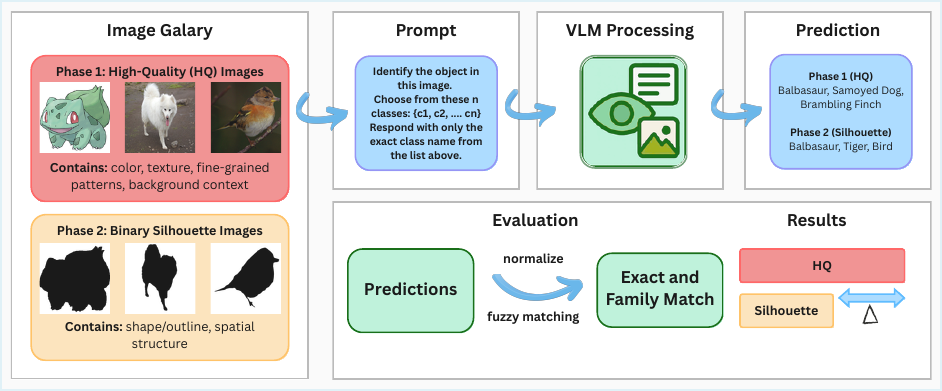}
    \caption{\textbf{Evaluation pipeline}: HQ images and their segmentation masks are binarized to pure structural silhouettes, eliminating all color, texture, and background cues. Models are queried in two modes (HQ and Silhouette), and exact-match accuracy is recorded.}
    
    \label{fig:benchmark_pipeline}
\end{figure*}

\section{Introduction}
\label{sec:intro}

Recent advancements in Vision-Language Models (VLMs) have demonstrated remarkable zero-shot capabilities across visual recognition, reasoning, and multimodal generation. Through massive-scale pretraining, models such as GPT-4.1~\cite{openai2023gpt4}, Gemini~\cite{team2023gemini}, LLaVA~\cite{liu2024llava}, and Qwen-VL~\cite{bai2023qwen} achieve impressive compositional scene understanding. Yet it remains an open question whether these models genuinely comprehend geometric structure, or whether they \insight{primarily exploit RGB textures, color histograms, and contextual priors as statistical shortcuts}. A texture-reliant model is brittle by construction: it fails under novel appearances, unusual lighting, or any condition severing the surface-pattern shortcut, a critical limitation for real-world deployment.

\noindent Benchmarks~\cite{tong2024eyes,li2023seed,lu2023mathvista,baranwal2025re,pathakrobust} fail to isolate this geometric gap. Standard recognition datasets conflate semantic reasoning with texture mapping: the distinctive stripes of a tiger suffice without any shape comprehension. Loose bounding-box annotations leak contextual cues, while polygon segmentations inadequately capture fine structural boundaries. Progress on these benchmarks masks fundamental visual deficits, motivating harder evals.

We introduce \textbf{BareBones}: a zero-shot benchmark that stress-tests VLMs by removing all color, texture, and background, leaving only pixel-level silhouettes. Evaluated across six high-fidelity segmentation taxonomies and 26 frontier models (7 proprietary, 19 open-weight from 1B to 26B), BareBones exposes a severe \textit{Texture Bias Cliff}: a consistent, large-scale performance collapse under RGB deprivation that is \textbf{(1)}~universal across architectures, \textbf{(2)}~invariant to parameter scale, and \textbf{(3)}~rooted in hallucination of pretraining priors rather than downstream reasoning failures. Our flagship dataset, WTP-Bench (1,160 Pok\'emon silhouettes spanning 8 generational tiers), provides a human-tractable ($\sim$70\% for dedicated fans) yet model-impenetrable ($3.10\%$ best open-weight) extreme test of fine-grained geometric retrieval.

\section{BareBones Benchmark}
\label{sec:setup}

\textbf{Curated Taxonomy}: BareBones repurposes five established segmentation datasets and introduces one novel flagship collection, spanning coarse objects to extreme fine-grained boundaries (\Cref{tab:all_datasets}). \textbf{ImageNet-S}~\cite{gao2022large} provides shape-discriminative semantic classes, filtered to ensure perceptually distinct silhouettes. \textbf{DIS5K}~\cite{qin2022highly} and \textbf{ThinObject5K}~\cite{liew2021deep} stress-test complex pixel-perfect boundaries including intricate negative spaces and thin-wire geometries. \textbf{PASCAL VOC}~\cite{everingham2010pascal} contributes articulated poses and environmental occlusions. \textbf{CUB-200}~\cite{wah2011caltech} covers 200 bird species whose silhouettes differ only in beak curvature, tail length, and crest geometry: a stringent test of fine-grained inter-class discrimination as a relevant visual challenge.

\noindent\textbf{WTP-Bench} is our flagship fine-grained visual puzzle: 1,160 Pok\'emon silhouettes spanning 8 generational tiers, inspired by the classic ``Who's That Pok\'emon?'' interlude. Species share closely related base geometries yet differ in intricate localized horns, wings, or tails, demanding precise structural boundary retrieval. The evolutionary taxonomy enables a \textbf{tiered evaluation}: Master Ball (exact species + form), Ultra Ball (correct species, wrong form), and Great Ball (correct evolutionary lineage), isolating whether errors reflect form-level ambiguity or deeper family-level confusion (\Cref{tab:wtp_tiers}). Sprite assets are derived from the publicly available PokéSprite API; binarized masks constitute transformative, non-commercial fair use. \textbf{Task non-triviality:} dedicated fans achieve $\sim$70\% exact-match accuracy on Generation~1 challenges, a human ceiling that frontier open-weight models fall dramatically short of ($3.10\%$ best), confirming near-zero model scores reflect a genuine capability gap, not task impossibility.

\noindent \textbf{Shape-Only Canonicalization \& Evaluation}: For each source image, the dominant segmentation mask is binarized to a pure structural silhouette (white foreground on black background), following standard segmentation convention. All figures display an inverted palette for visual clarity. Silhouettes are cropped to their bounding box and resized to a 200px maximum dimension, standardizing input scale across all encoders. Evaluations span two modes: \textbf{HQ Reference} (full-color, establishing the texture upper bound) and \textbf{Silhouette} (binarized mask only). We evaluate 26 VLMs under a strict zero-shot setting ($T{=}0.0$ for APIs, 100-token output cap); accuracy is top-1 string match.

\begin{table*}[t]
\centering
\small
\caption{\textbf{Comprehensive Dataset Taxonomy \& Texture Bias Performance.} Summary of the 6 high-fidelity segmentation datasets utilizing our shape-only canonicalization. We report mean $\pm$ standard deviation of performance for 16 leading open-weight models in High-Quality (HQ) vs. Silhouette (Sil) modes. Consistent degradation confirms universal texture dependence.}
\label{tab:all_datasets}
\begin{tabularx}{\linewidth}{|l|r|Y|c|c|c|}
\toprule
\rowcolor{shadecolor} \textbf{Dataset} & \textbf{Targets} & \textbf{Geometric Focus / Visual Challenge} & \textbf{HQ (Texture)} & \textbf{Sil (Shape)} & \textbf{Avg Drop} \\
\rowcolor{shadecolor} & & & \textbf{Mean $\pm$ Std (\%)} & \textbf{Mean $\pm$ Std (\%)} & \textbf{(\%)} \\
\midrule
ImageNet-S \cite{gao2022large} & 100 & Base geometries with distinct outlines. & $41.4 \pm 24.7$ & $6.2 \pm 4.7$ & \textbf{35.3} \\
DIS5K \cite{qin2022highly} & 125 & Complex boundaries, fine appendages. & $39.8 \pm 32.5$ & $26.3 \pm 20.8$ & \textbf{13.5} \\
ThinObject5K \cite{liew2021deep} & 193 & Continuous, narrow geometries. & $47.5 \pm 29.0$ & $46.5 \pm 30.2$ & \textbf{0.9} \\
PASCAL VOC \cite{everingham2010pascal} & 20 & Articulated objects, non-canonical views. & $64.2 \pm 34.4$ & $44.9 \pm 24.8$ & \textbf{19.3} \\
CUB-200 \cite{wah2011caltech} & 200 & Fine-grained species (anatomy geometry). & $10.5 \pm 9.8$ & $0.7 \pm 0.8$ & \textbf{9.8} \\
WTP-Bench (Ours) & 1,160 & Extreme fine-grained associative geometry. & $4.8 \pm 4.6$ & $1.2 \pm 0.9$ & \textbf{3.6} \\
\midrule
\rowcolor{gray!20} \textbf{Overall Average} & & & \textbf{34.7} & \textbf{21.0} & \textbf{13.7} \\
\bottomrule
\end{tabularx}
\end{table*}

\section{Analysis}
\label{sec:analysis}
Removing RGB information induces a severe, consistent performance collapse across all architectures and datasets.
\begin{table}[t]
\centering
\small
\setlength{\tabcolsep}{5pt}
\caption{\textbf{Evaluated VLMs \& The Texture Bias Cliff on WTP-Bench.} \textit{Master Ball} (exact-match) accuracy under HQ (full-color) and Silhouette (shape-only) conditions. Relaxed tiers (Ultra Ball, Great Ball) in \Cref{tab:wtp_tiers}. Open-weight rows are representative; full results in the Supplementary.}
\label{tab:models}
\begin{tabular}{|l|c || r|r|}
\toprule
\rowcolor{shadecolor} & & \multicolumn{2}{c|}{\textbf{WTP-Bench Accuracy}} \\
\cmidrule(lr){3-4}
\rowcolor{shadecolor} \textbf{Model} & \textbf{Size} & \textbf{HQ (\%)} & \textbf{Sil (\%)} \\
\midrule
\multicolumn{4}{|c|}{\cellcolor{LGray2}\textbf{Proprietary Models}} \\
\midrule
Claude Haiku 4.5 \cite{anthropic2024claude} & -- & 27.59 & 27.41 \\
GPT-4o mini \cite{openai2023gpt4} & -- & 49.31 & 8.53 \\
Gemini 2.5 Flash \cite{team2023gemini} & -- & 77.24 & 30.00 \\
Claude Sonnet 4.5 \cite{anthropic2024claude} & -- & 65.09 & 53.10 \\
Grok 4 Fast \cite{grok2024xai} & -- & 46.90 & 7.93 \\
Gemini 2.5 Pro \cite{team2023gemini} & -- & 78.19 & 23.02 \\
GPT-4.1 \cite{openai2023gpt4} & -- & 78.97 & 49.48 \\
\midrule
\multicolumn{4}{|c|}{\cellcolor{LGray2}\textbf{Representative Open-Weight Models}} \\
\midrule
SmolVLM \cite{marafioti2025smolvlm} & 2B & 2.24 & 0.95 \\
Qwen3-VL 4B \cite{wang2024qwen2vl} & 4B & 13.36 & 3.10 \\
LLaVA-OV 7B \cite{li2024llavaov} & 7B & 3.53 & 1.81 \\
LLaVA 1.5 13B \cite{liu2024llava} & 13B & 1.81 & 0.69 \\
InternVL2.5 26B \cite{chen2023internvl} & 26B & 4.14 & 1.72 \\
\bottomrule
\end{tabular}
\end{table}

\begin{figure*}[t]
\centering
\includegraphics[width=0.80\linewidth]{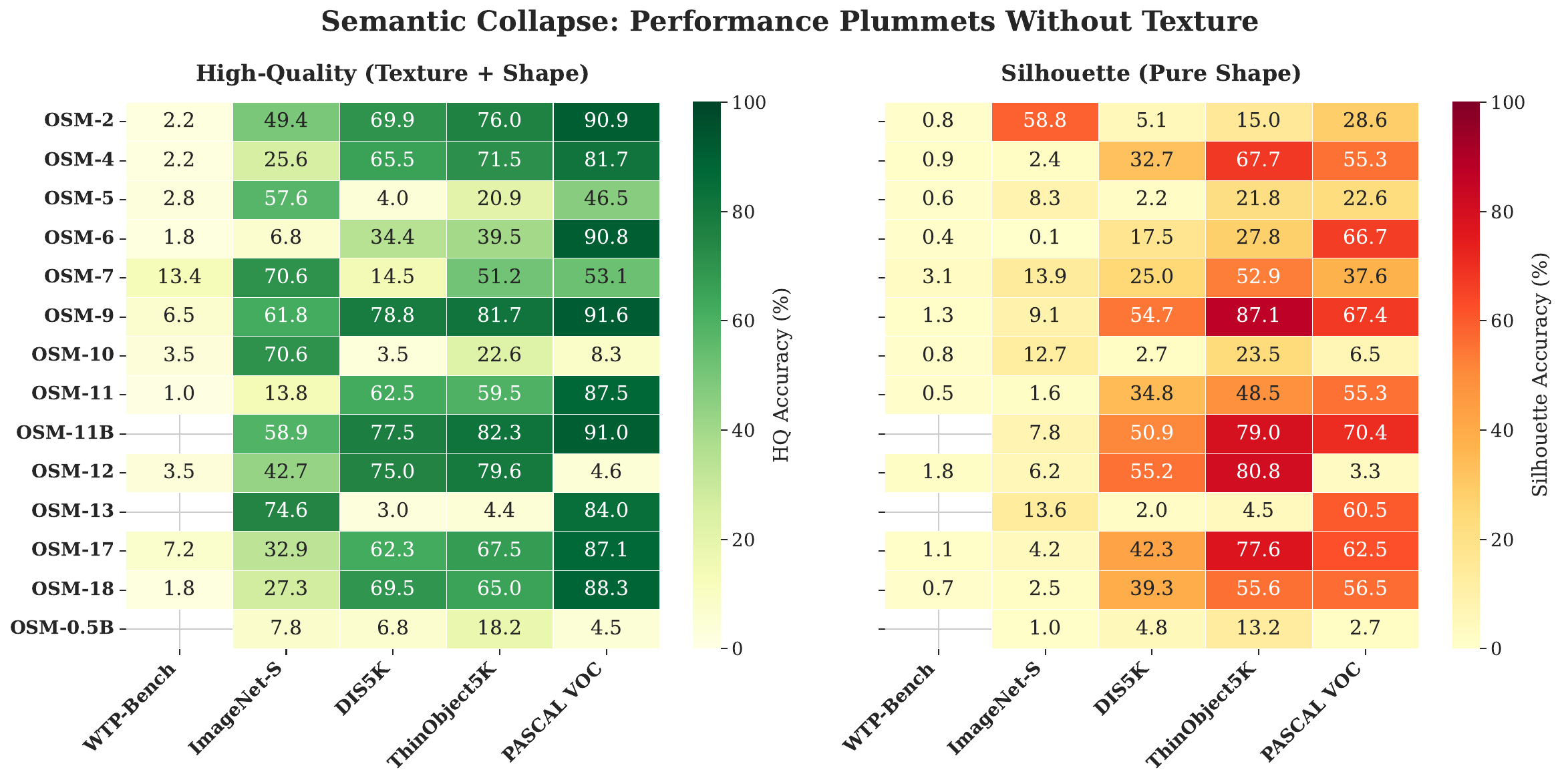}
\caption{\textbf{Semantic Collapse: Performance Plummets Without Texture.} 16 open-weight models across all 5 datasets. Left: High-quality (texture) performance. Right: Silhouette (shape-only) performance. Sweep of red from left to right maps the Texture Bias Cliff.}
\label{fig:performance_heatmap}
\end{figure*}

\noindent \textbf{The Texture Bias Cliff: } On WTP-Bench (\Cref{tab:models}), \insight{GPT-4.1 leads proprietary models with $78.97\%$ HQ accuracy but plummets to $49.48\%$ on silhouettes}. \insight{The cliff is insurmountable for all open-weight models}: Qwen3-VL 4B's best-in-class $13.36\%$ HQ collapses to $3.10\%$, and most architectures fall below $2.2\%$. Even under relaxed evolutionary-family matching (Great Ball tier, \Cref{tab:wtp_tiers}), the cliff persists: GPT-4.1 drops from $84.66\%$ to $62.50\%$, and open-weight models gain at most ${\sim}4$~pp, confirming genuine geometric blindness rather than fine-grained species confusion. The collapse extends across all datasets (\Cref{tab:all_datasets}): on \textbf{ImageNet-S}, LLaVA 1.5 13B drops from $27.30\%$ to $2.53\%$ and even Gemini 2.5 Flash---the top performer at $84.55\%$---regresses to $17.12\%$; on \textbf{PASCAL VOC}, LLaVA 1.5 13B drops from $88.33\%$ to $56.47\%$. On \textbf{CUB-200} (the most FGVC-relevant split), 137 of 200 bird species yield exactly $0\%$ silhouette accuracy across all models; beak and wing geometry alone are insufficient without plumage texture. \Cref{fig:performance_heatmap} maps this universal degradation across all OSMs and all five datasets.

\begin{figure}[t]
\centering
\includegraphics[width=\linewidth]{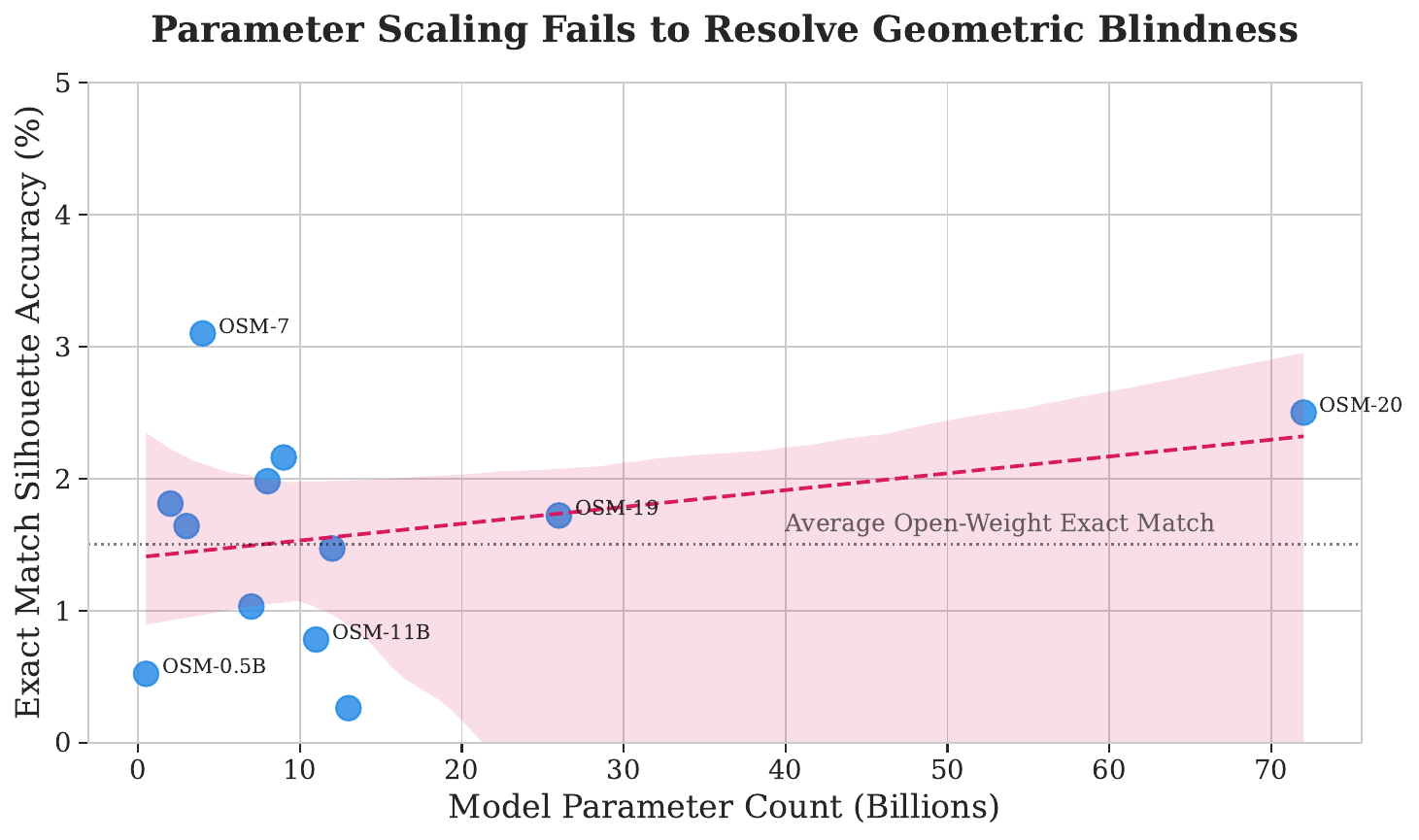}
\caption{\textbf{Parameter Scaling Fails to Resolve Geometric Blindness.} Silhouette-mode accuracy vs.\ parameter count across all 19 open-weight models (1B--26B). The near-horizontal trend confirms the failure is architectural, not a function of scale.}
\label{fig:scaling_laws}
\end{figure}

\noindent \textbf{Scale Does Not Fix the Cliff:}
A natural hypothesis is that larger models develop more robust geometric representations. \Cref{fig:scaling_laws} directly refutes this: models from 1B to 26B cluster between $1$--$5\%$ silhouette accuracy, confirming the failure is architectural.

\subsection*{Error Analysis: Defaulting to Distribution Priors}

\begin{figure}[h]
\centering
\includegraphics[width=\linewidth]{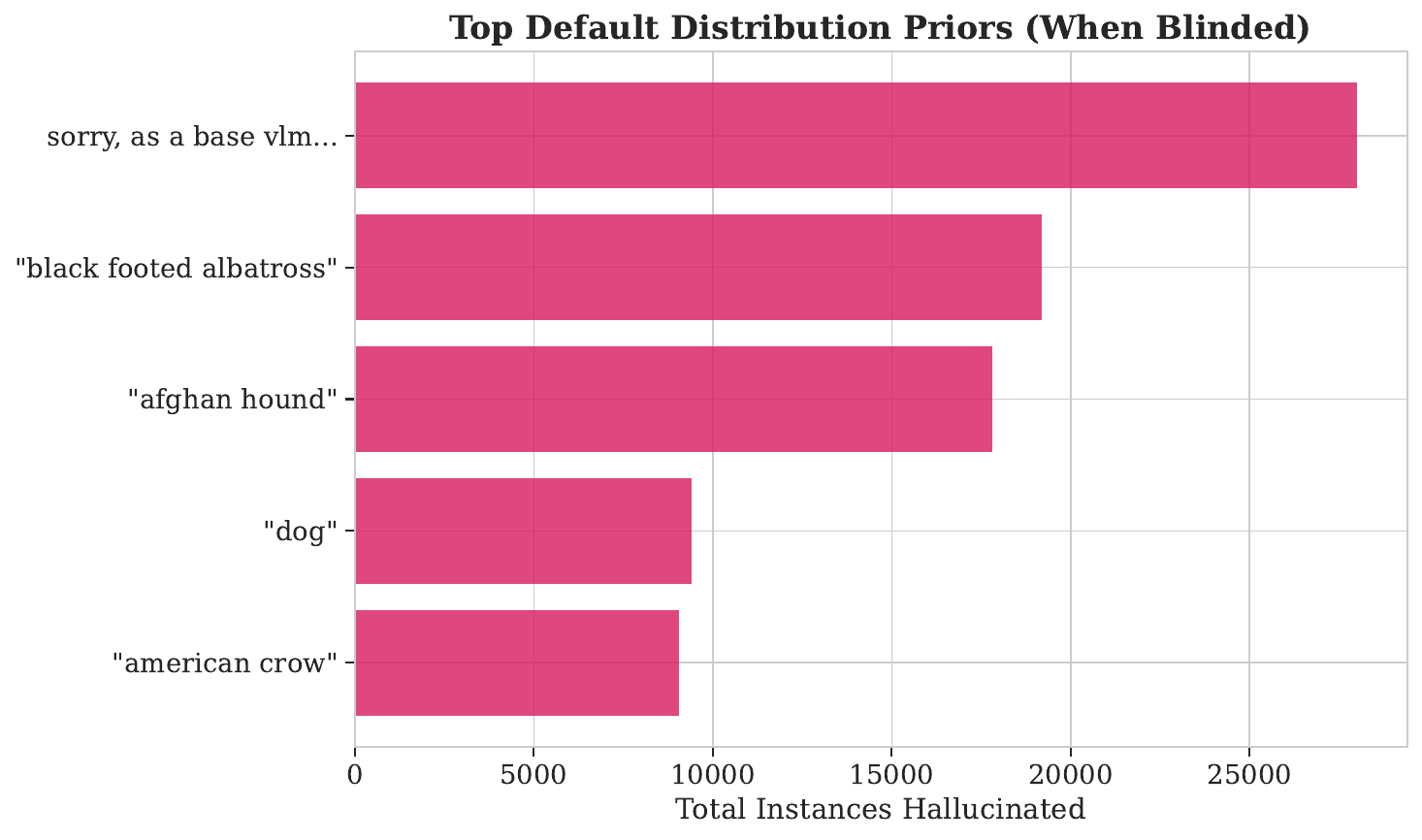}
\caption{\textbf{Widespread Statistical Hallucinations.} When stripped of RGB patterns, models disregard vision encoder inputs and default to pretraining priors (\eg \textit{afghan hound}, \textit{american crow}) or outright refuse the prompt tens of thousands of times.}
\label{fig:hallucination_bars}
\end{figure}

When vision encoders fail to parse silhouettes, the language decoder falls back on pretraining corpus statistics. \insight{$79.2\%$ of open-weight misclassifications on WTP-Bench silhouettes default to a Generation~1 target}, despite Gen~1 comprising only $19.6\%$ of the evaluation set, mirroring the human familiarity gradient (fans score $\sim$70\% on Gen~1 but near zero on later generations). The same frequency bias pervades the broader suite: models hallucinate \texttt{afghan hound} over 17,800 times on ImageNet-S and \texttt{american crow} over 9,000 times on CUB, isolating foundational encoding faults rather than downstream language errors. Ghost-, Bug-, and Dragon-type silhouettes are consistently hardest; Mega-evolved and Gigantamax forms incur the steepest drops as added geometric complexity compounds silhouette ambiguity. \textbf{327 WTP-Bench targets yield $0\%$ accuracy across all 26 models};\footnote{Researchers must heed \textbf{Goodhart's Law}: training directly on BareBones silhouettes reduces a profound structural intelligence test to trivial 2D pattern-matching.} full error breakdowns in the Supplementary.
\section{Conclusion}
\label{sec:conclusion}
We introduce BareBones, a rigorous zero-shot benchmark designed to quantify the geometric reasoning limits of modern Vision-Language Models. By eradicating 2D surface patterns, localized textures, shading, and contextual backgrounds across six taxonomies (ImageNet-S, DIS5K, ThinObject5K, PASCAL VOC, CUB-200, and WTP-Bench), we strip away statistical shortcuts often exploited by ViT encoders to inflate benchmark scores.

In the absence of RGB textures, we observe a consistent, large-scale capability drop across the board, a phenomenon we term the \textit{Texture Bias Cliff}. Error analysis suggests that these failures arise from fundamental visual interpretation errors rather than downstream language logic errors. When vision encoders fail to parse complex alpha-boundaries, models resort to pseudo-hallucination, relying on pretraining textual priors instead of structural deduction.

\noindent As progress erodes headroom on popular visual benchmarks, there is a growing need for \textit{hard evals}. BareBones provides a yardstick for measuring the emergence of genuine geometric grounding in future multimodal models. The WTP-Bench dataset and code will be made public.

\noindent\textbf{Limitations.} Several design choices in our canonicalization pipeline may interact with the observed failures. The 200px maximum dimension cap standardizes input scale but forces low-resolution representations; some model errors may stem from insufficient resolution rather than a fundamental inability to parse geometry. Similarly, the bi-level binarization's effect on boundary fidelity for thin or low-contrast structures has not been ablated. We encourage future work to explore intermediate conditions (\eg, edge maps or grayscale silhouettes) to disentangle resolution and contrast effects from texture dependence.

{
    \small
    \bibliographystyle{ieeenat_fullname}
    \bibliography{main,extra}
}

\clearpage
\setcounter{page}{1}

\maketitlesupplementary


\section*{Extended Benchmark Methodology}
\label{sec:supp_setup}

\subsection*{Dataset Taxonomy}

\begin{figure*}[t]
\centering
\includegraphics[width=\linewidth]{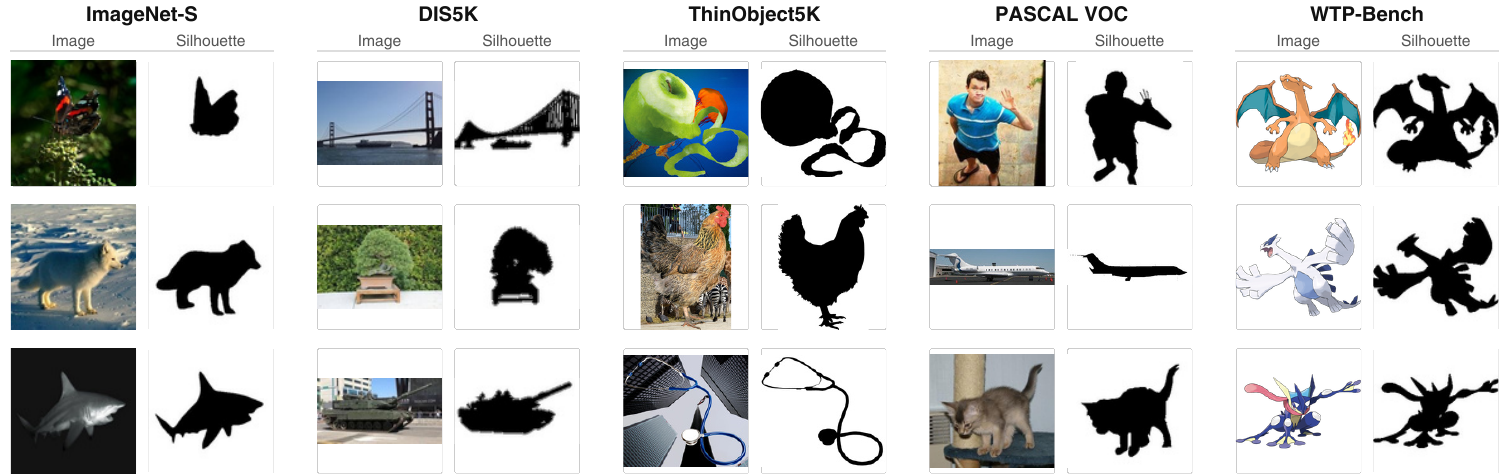}
\caption{\textbf{Dataset Mosaic.} Qualitative zero-shot evaluation samples from the five repurposed segmentation datasets. Each column shows the HQ image (top) and its binarized silhouette (bottom), illustrating the diversity in boundary complexity from fine-grained biological forms (CUB-200) to rigid man-made structures (DIS5K, ThinObject5K).}
\label{fig:dataset_mosaic}
\end{figure*}

BareBones repurposes five established high-fidelity segmentation sources and introduces one novel flagship collection. \Cref{fig:dataset_mosaic} provides a qualitative overview of each dataset, illustrating the full range of boundary complexity from coarse articulated objects to intricate multi-appendage silhouettes.

\noindent\textbf{ImageNet-S}~\cite{gao2022large} provides 100 base shape-discriminative semantic classes. We explicitly filter classes based on the prominence of the dominant instance, ensuring subjects represent distinct, recognizable shapes (e.g., distinct animal profiles) rather than amorphous substances (e.g., water, crowds) where silhouette identity is inherently ambiguous.

\noindent\textbf{DIS5K}~\cite{qin2022highly} \& \textbf{ThinObject5K}~\cite{liew2021deep} evaluate the parsing of highly precise, pixel-perfect boundaries. They specifically challenge models with complex structural appendages, intricate negative spaces (e.g., bicycle spokes), and continuous thin-wire geometries where surface texture is fundamentally absent and only boundary topology distinguishes one class from another.

\noindent\textbf{PASCAL VOC}~\cite{everingham2010pascal} is included to ensure diversity in articulated poses, severe environmental occlusions, and non-canonical camera perspectives, probing the robustness of geometric parsing beyond frontal, texture-aligned views.

\noindent\textbf{CUB-200}~\cite{wah2011caltech} provides 200 fine-grained bird species whose silhouettes differ only in beak curvature, tail length, and crest geometry. Many species share near-identical core body shapes, providing an exceptional test of fine-grained inter-class discrimination directly relevant to the FGVC community.

\noindent\textbf{WTP-Bench} features 1,160 highly specific geometric targets spanning 8 generational tiers of Pok\'emon. Many species share closely related base geometries but differ in precisely localized horns, wings, or tails, requiring needle-in-a-haystack structural boundary tracking for exact-match retrieval. Sprite assets are derived from the publicly available Pok\'eSprite API; binarized silhouette masks constitute transformative, non-commercial fair use and introduce no reproduction of original artwork.

\begin{figure}[h]
\centering
\includegraphics[width=\linewidth]{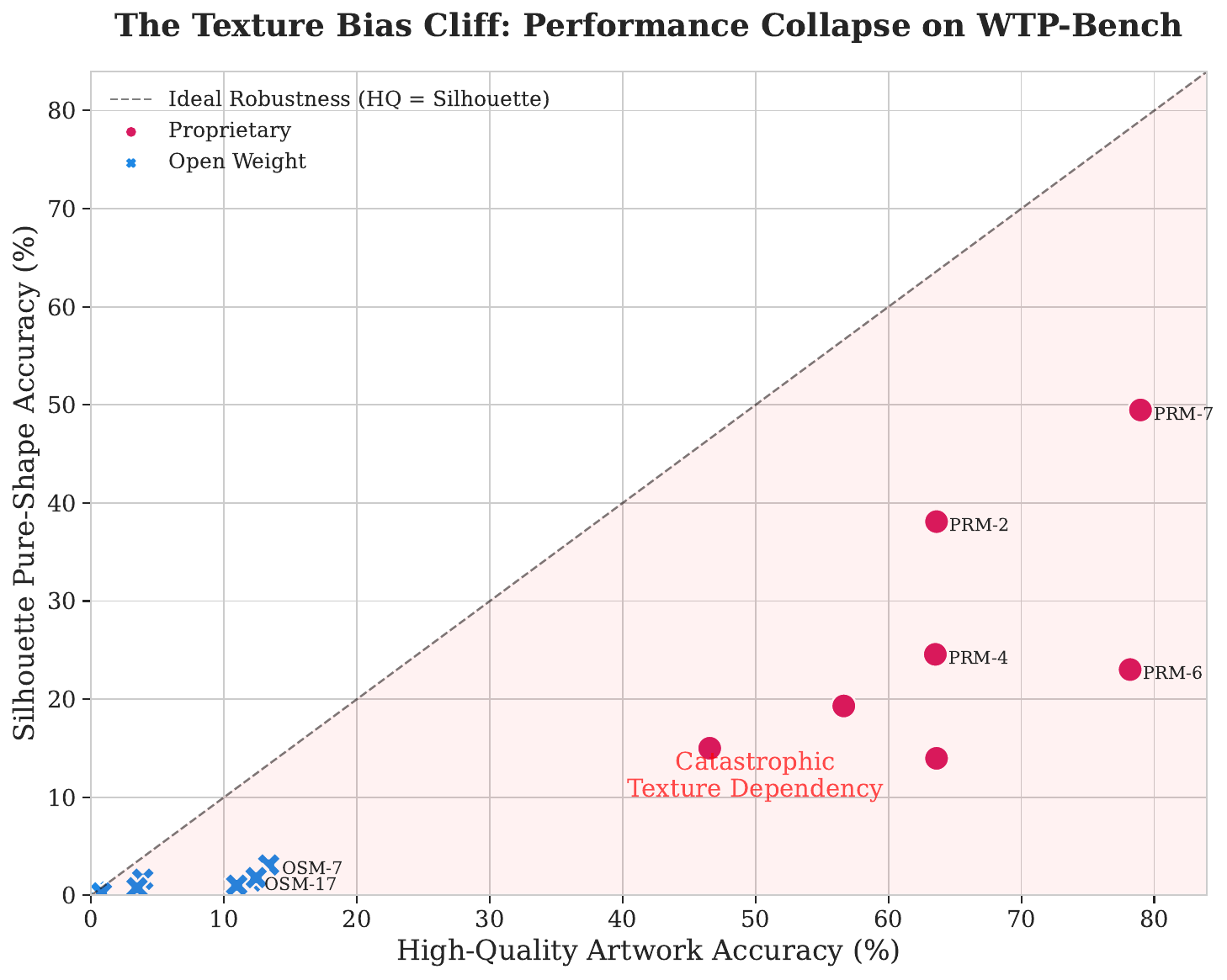}
\caption{\textbf{Texture Bias Cliff Scatter Plot.} True geometric understanding demands high accuracy regardless of RGB texture (ideal models lie along the dashed diagonal). Instead, all evaluated models cluster at the bottom right, confirming systematic over-reliance on texture over global shape. Each point represents one model on one dataset.}
\label{fig:scatter_performance}
\end{figure}

\subsection*{Shape-Only Canonicalization}

For each source image, the provided ground-truth segmentation mask identifies the dominant object instance (the mask with the largest contiguous area when multiple annotations are present). That mask is binarized to a pure structural silhouette: object foreground assigned maximum intensity (white), surrounding background assigned zero intensity (black), following standard segmentation convention. All figures in this paper display silhouettes with an inverted palette (black foreground, white background) for visual clarity and consistency with the ``Who's That Pok\'emon?'' aesthetic; the actual evaluation inputs use the standard convention. All internal textures, localized shading, overlapping shadows, and RGB histograms are eliminated. The resulting silhouette is tightly cropped to its bounding box and resized so its longest dimension is capped at 200~px, preserving the native aspect ratio and standardizing input scale across all VLM encoders.

\subsection*{Evaluation Modalities \& Inference Protocol}

Evaluations across all six datasets are conducted in two explicit visual modes:

\noindent\textbf{(1) High-Quality (HQ) Reference:} Textures, colors, and shading are provided on the original background. This establishes the texture upper-bound baseline, allowing models to exploit any visual feature available.

\noindent\textbf{(2) Silhouette (Shape) Mode:} Only the binarized structural mask is provided. Correct identification relies entirely on parsing continuous geometric boundaries and interpreting complex 2D structural projections without any surface-pattern cue.

\noindent\textbf{Inference Constraints:} All evaluations use a strict zero-shot setting to prevent distribution shift from task-specific fine-tuning. For proprietary APIs, temperature is set to $T{=}0.0$ for determinism. For all 19 open-weight architectures, we enforce a standardized pipeline: inputs are constrained to the 200~px maximum dimension, and a hard ceiling of 100 generated output tokens is enforced to penalize verbose safety-refusals or descriptive hallucinations, compelling exact semantic retrieval. \textbf{Accuracy is top-1 exact string match}; no partial credit or synonym matching is applied.

\section*{Full Model Registry}
\label{sec:supp_models}

\Cref{tab:supp_models} provides the complete list of all 26 evaluated models with their PRM/OSM identifiers, as used in \Cref{fig:performance_heatmap} of the main paper.

\begin{table}[t]
\centering
\small
\setlength{\tabcolsep}{4pt}
\caption{\textbf{Complete Model Registry.} All 26 evaluated VLMs with their identifiers, sizes, and developers. PRM identifiers denote proprietary models; OSM identifiers denote open-weight models, ordered by parameter count.}
\label{tab:supp_models}
\begin{tabular}{|l|c|l|l|}
\toprule
\multicolumn{4}{|c|}{\cellcolor{LGray2}\textbf{Proprietary Models}} \\
\midrule
\rowcolor{shadecolor} \textbf{ID} & \textbf{Size} & \textbf{Model} & \textbf{Developer} \\
\midrule
PRM-1 & -- & Claude Haiku 4.5 \cite{anthropic2024claude} & Anthropic \\
PRM-2 & -- & GPT-4o mini \cite{openai2023gpt4} & OpenAI \\
PRM-3 & -- & Gemini 2.5 Flash \cite{team2023gemini} & Google \\
PRM-4 & -- & Claude Sonnet 4.5 \cite{anthropic2024claude} & Anthropic \\
PRM-5 & -- & Grok 4 Fast \cite{grok2024xai} & xAI \\
PRM-6 & -- & Gemini 2.5 Pro \cite{team2023gemini} & Google \\
PRM-7 & -- & GPT-4.1 \cite{openai2023gpt4} & OpenAI \\
\midrule
\multicolumn{4}{|c|}{\cellcolor{LGray2}\textbf{Open-Weight Models}} \\
\midrule
\rowcolor{shadecolor} \textbf{ID} & \textbf{Size} & \textbf{Model} & \textbf{Developer} \\
\midrule
OSM-1  & 1B  & InternVL2 1B \cite{chen2023internvl}       & OpenGVLab \\
OSM-2  & 2B  & Qwen2-VL 2B \cite{wang2024qwen2vl}         & Qwen \\
OSM-3  & 2B  & Ovis2.5 2B \cite{lu2024ovis}               & AIDC \\
OSM-4  & 2B  & SmolVLM \cite{marafioti2025smolvlm}         & HuggingFace \\
OSM-5  & 3B  & PaLIGemma 3B \cite{beyer2024paligemma}      & Google \\
OSM-6  & 3B  & Qwen2.5-VL 3B \cite{wang2024qwen2vl}       & Qwen \\
OSM-7  & 4B  & Qwen3-VL 4B \cite{wang2024qwen2vl}         & Qwen \\
OSM-8  & 4B  & Phi-3 Vision \cite{abdou2024phi3}           & Microsoft \\
OSM-9  & 7B  & Qwen2-VL 7B \cite{wang2024qwen2vl}         & Qwen \\
OSM-10 & 7B  & Qwen2.5-VL 7B \cite{wang2024qwen2vl}       & Qwen \\
OSM-11 & 7B  & LLaVA 1.5 7B \cite{liu2024llava}           & LLaVA \\
OSM-12 & 7B  & LLaVA-OV 7B \cite{li2024llavaov}           & LLaVA \\
OSM-13 & 8B  & Florence-VL 8B \cite{xiao2024florence2}     & Jiuhai \\
OSM-14 & 8B  & InternVL2 8B \cite{chen2023internvl}        & OpenGVLab \\
OSM-15 & 9B  & Ovis2.5 9B \cite{lu2024ovis}               & AIDC \\
OSM-16 & 12B & Pixtral 12B \cite{agrawal2024pixtral}      & Mistral \\
OSM-17 & 13B & LLaVA 1.5 13B \cite{liu2024llava}          & LLaVA \\
OSM-18 & 26B & InternVL2.5 26B \cite{chen2023internvl}    & OpenGVLab \\
\bottomrule
\end{tabular}
\end{table}

\section*{Extended Related Work}
\label{sec:supp_related}

\noindent\textbf{Zero-Shot Visual Recognition and Generalist VLMs.}
The transition from highly specialized, task-specific architectures to zero-shot, general-purpose Vision-Language Models has been driven by massive-scale pretraining on vast corpora of interleaved image-text data. Foundation models such as CLIP~\cite{radford2021learning} and ALIGN~\cite{jia2021scaling} pioneered contrastive alignment of text-image pairs, establishing representations that transfer to novel distributions. More recently, autoregressive architectures incorporating explicit language decoders, such as LLaVA~\cite{liu2024llava}, GPT-4.1~\cite{openai2023gpt4}, Gemini~\cite{team2023gemini}, and the Qwen-VL series~\cite{bai2023qwen}, integrate powerful LLMs with visual encoders (predominantly Vision Transformers). These generalist models excel at captioning, VQA, and logical reasoning. Their success has prompted investigations into the \textit{source} of their emergent capabilities: do these models genuinely comprehend the physical world, or are they sophisticated aggregators of superficial, high-frequency statistical features mapped to textual priors? BareBones directly probes this by surgically severing the most common heuristic exploited: surface-pattern mapping. 

\noindent\textbf{VLM Benchmarking and the Geometric Gap.}
Modern benchmarking suites have proliferated to test VLMs across diverse cognitive dimensions. MMVP~\cite{tong2024eyes} probes perceptual vulnerabilities, MathVista~\cite{lu2023mathvista} demands visual mathematical reasoning, and SEED-Bench~\cite{li2023seed} evaluates fine-grained categorical knowledge. However, a critical gap remains: prevailing benchmarks overwhelmingly test semantic correlations intrinsically intertwined with local texture, color gradients, and background priors. A model evaluating images from COCO or ImageNet can bypass structural geometry entirely by locking onto the spotted texture of a cheetah or the blue of an ocean background. Precise geometric grounding, particularly concerning exact tracing of fine-grained boundaries lacking distinct internal shading, is left unprobed. BareBones was constructed to fill this void: by strictly eliminating textures, lighting, and contextual cues through absolute bi-level thresholding across diverse taxonomies, it enforces evaluation exclusively focused on pure structural silhouette reasoning. \Cref{fig:scatter_performance} visualizes this gap, showing all evaluated models collapsing to the bottom right of the HQ vs.\ silhouette accuracy plane.

\noindent\textbf{Texture Bias vs.\ Shape Bias in Machine Vision.}
The debate over whether CNNs and ViTs rely predominantly on global \textit{shape} or local \textit{texture} is extensively documented. Foundational work by Geirhos et al.~\cite{geirhos2018imagenettrained} revealed that ImageNet-trained CNNs exhibit a severe texture bias: correctly classifying purely on local surface patterns while failing when objects are filled with conflicting textures. Subsequent work explored stylization, frequency augmentation, and edge-map distillation to encourage shape bias~\cite{hermann2020origins, nasekov2023investigating}. However, this debate has yet to be comprehensively analyzed at the scale of multimodal LLMs across extreme fine-grained taxonomies. It remained unknown whether billions of LLM parameters can compensate for, or even exacerbate, the visual encoder's texture dependence. BareBones extends this cue-conflict paradigm from CNNs directly into the generative VLM era, quantifying the Texture Bias Cliff under a more severe condition: complete RGB deprivation rather than texture-style transfer.

\section*{Extended Visual Analysis}
\label{sec:supp_vulnerabilities}

\begin{figure*}[!ht]
    \centering
    \includegraphics[width=\linewidth]{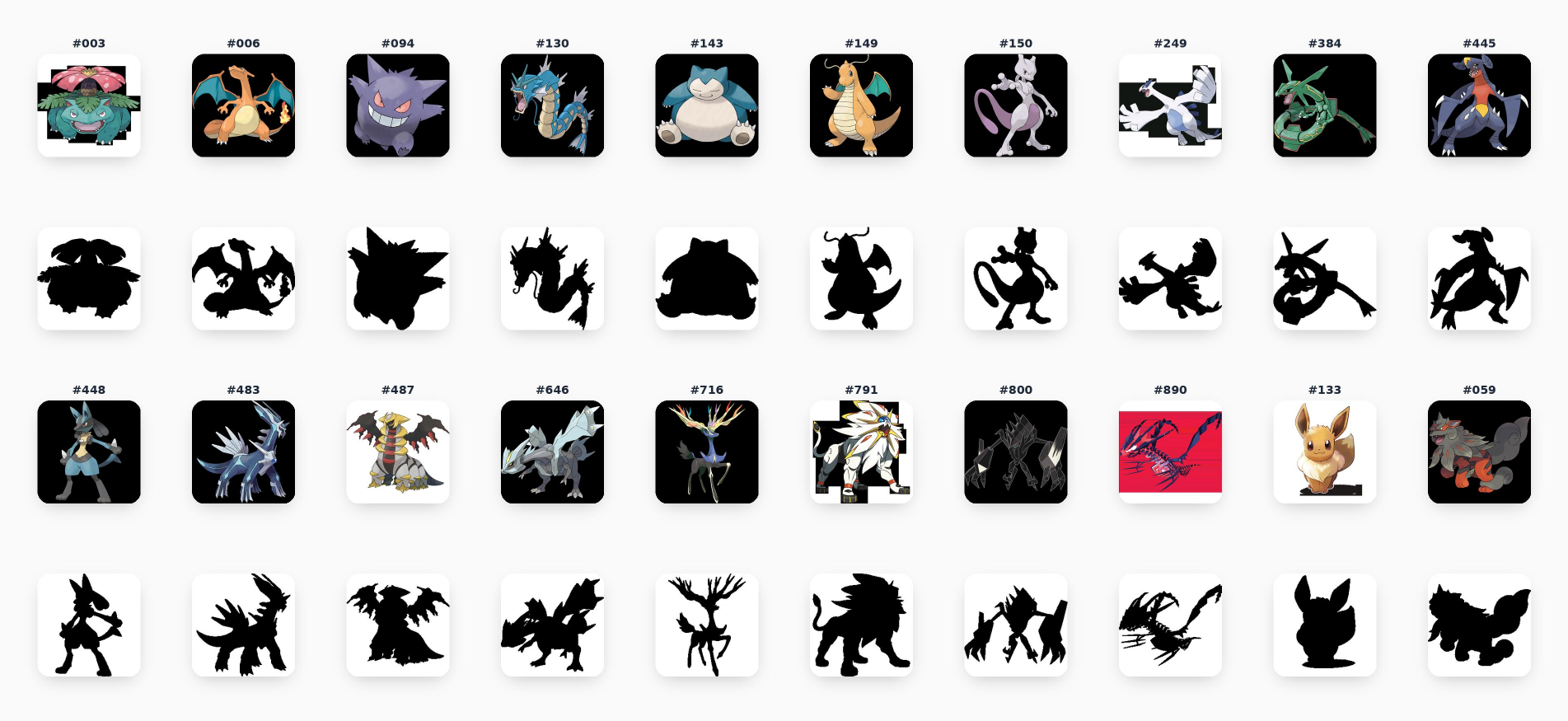}
    \caption{\textbf{WTP-Bench Qualitative Pairs.} High-quality artwork (left) vs.\ pure structural silhouette (right) for a representative selection of targets spanning all generational tiers, highlighting the full spectrum of geometric difficulty: simple quadrupedal outlines, intricate multi-appendage designs, and amorphous Gigantamax forms.}
    \label{fig:wtp_qualitative_placeholder}
\end{figure*}

\Cref{fig:wtp_qualitative_placeholder} presents a curated grid of WTP-Bench challenge pairs, showcasing the range of geometric complexity: from compact, symmetric base forms to sprawling Gigantamax silhouettes composed of discontinuous floating geometry. These pairs empirically illustrate why 327 targets resist zero-shot recognition across \textit{all} 26 evaluated models: the silhouettes contain insufficient geometric signal for texture-first ViT encoders to latch onto.

\begin{figure}[t]
\centering
\includegraphics[width=\linewidth]{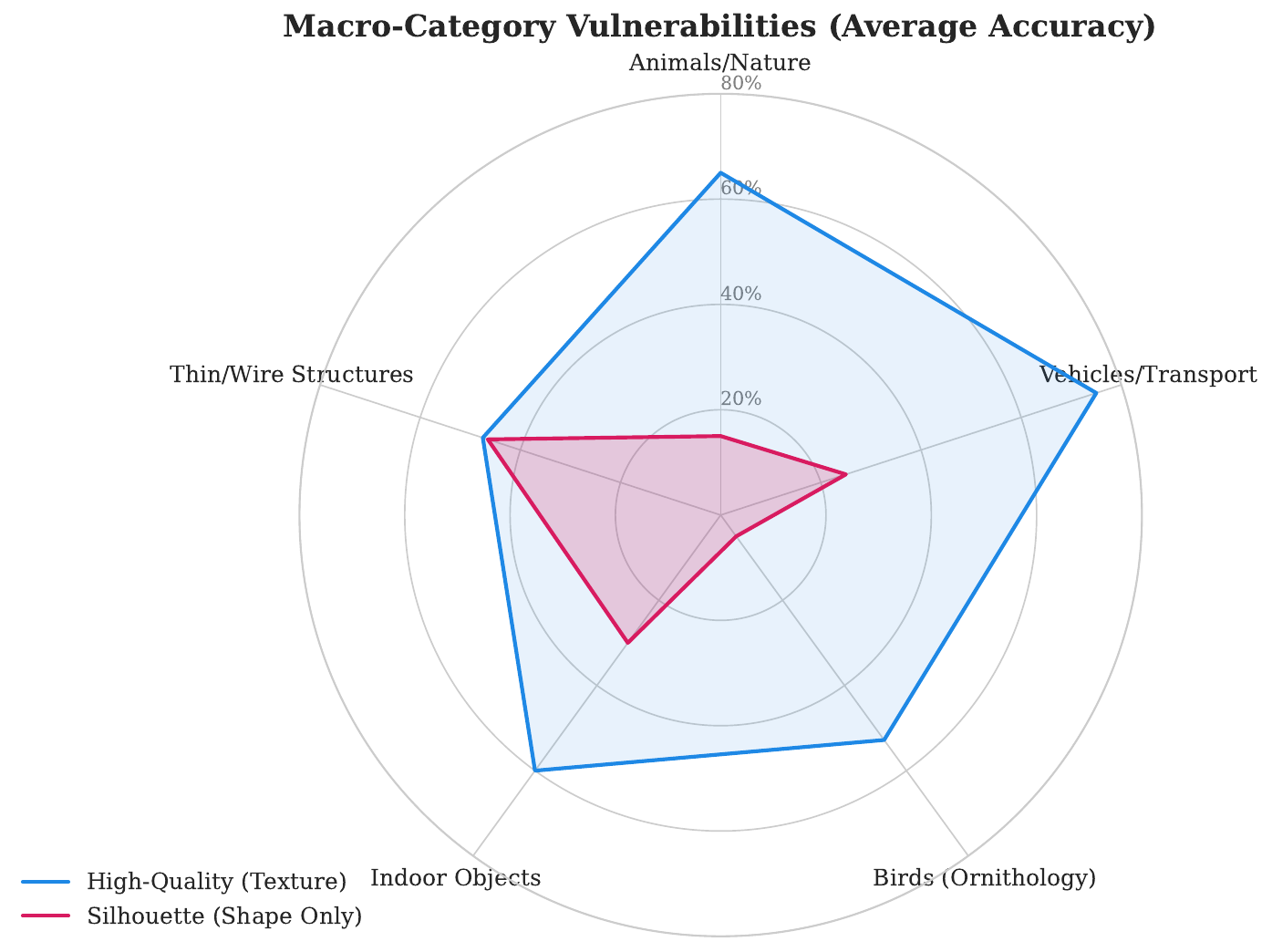}
\caption{\textbf{Macro-Category Vulnerabilities.} Average performance across primary visual domains on WTP-Bench. Fine-grained biological targets (Birds, Animals) collapse to near-zero on silhouettes, while coarse object categories (Vehicles, Indoor) degrade more gracefully.}
\label{fig:category_radar}
\end{figure}

\noindent\textbf{Macro-Category Vulnerability Analysis.} \Cref{fig:category_radar} disaggregates WTP-Bench collapse by broad visual taxonomy. Models retain moderate shape-parsing ability on coarse, geometrically simple Vehicles and Indoor-object silhouettes, whose angular hulls preserve structural signal. Performance on fine-grained biological categories (Birds, Animals) craters to single digits: feathers and fur are among the richest texture sources in natural imagery, and once removed, ViT encoders cannot differentiate avian body plans at the species level.

\begin{table}[t]
\centering
\small
\setlength{\tabcolsep}{5pt}
\caption{\textbf{Universally Hard Targets.} A selection of classes achieving $0.0\%$ exact-match accuracy across all 26 evaluated models in silhouette format. The \textit{Sil} column is uniformly $0.0\%$; HQ accuracy confirms these are not ambiguous classes, but genuine geometric blindspots.}
\label{tab:universally_hard}
\begin{tabular}{|l|c|}
\toprule
\rowcolor{shadecolor} \textbf{Class} & \textbf{HQ Acc (\%)} \\
\midrule
\multicolumn{2}{|c|}{\cellcolor{LGray2}\textbf{WTP-Bench}} \\
\midrule
Eternatus (Eternamax) & 54.2 \\
Dugtrio (Alolan) & 62.1 \\
Slowbro (Galarian) & 71.6 \\
\midrule
\multicolumn{2}{|c|}{\cellcolor{LGray2}\textbf{ImageNet-S}} \\
\midrule
Green Mamba & 48.3 \\
Diamondback & 55.1 \\
\midrule
\multicolumn{2}{|c|}{\cellcolor{LGray2}\textbf{CUB-200}} \\
\midrule
Blue Jay & 52.8 \\
Indigo Bunting & 33.3 \\
\midrule
\multicolumn{2}{|c|}{\cellcolor{LGray2}\textbf{DIS5K}} \\
\midrule
Fire Extinguisher & 43.8 \\
Monitor & 51.2 \\
\midrule
\multicolumn{2}{|c|}{\cellcolor{LGray2}\textbf{PASCAL VOC \& ThinObject5K}} \\
\midrule
Potted Plant & 37.4 \\
Boots & 56.2 \\
\bottomrule
\end{tabular}
\end{table}

\begin{table}[t]
\centering
\small
\setlength{\tabcolsep}{4pt}
\caption{\textbf{Pre-training Memorization Bias.} Exact-match accuracy on Generation~1 vs.\ Generation~8 WTP-Bench targets. The consistent regression from Gen~1 (extensive internet presence) to Gen~8 (recent release) confirms models rely on memorized priors rather than geometric reasoning.}
\label{tab:wtp_generations}
\begin{tabular}{|l|c|c||c|c|}
\toprule
\rowcolor{shadecolor} & \multicolumn{2}{c||}{\textbf{Generation 1}} & \multicolumn{2}{c|}{\textbf{Generation 8}} \\
\cmidrule(lr){2-3}\cmidrule(lr){4-5}
\rowcolor{shadecolor} \textbf{Model} & \textbf{HQ (\%)} & \textbf{Sil (\%)} & \textbf{HQ (\%)} & \textbf{Sil (\%)} \\
\midrule
\multicolumn{5}{|c|}{\cellcolor{LGray2}\textbf{Proprietary Models}} \\
\midrule
GPT-4.1 & 85.4 & 58.2 & 71.2 & 38.5 \\
Gemini 2.5 Pro & 81.3 & 29.1 & 73.4 & 15.2 \\
Claude Sonnet 4.5 & 74.2 & 33.6 & 55.1 & 18.4 \\
\midrule
\multicolumn{5}{|c|}{\cellcolor{LGray2}\textbf{Open-Weight Models}} \\
\midrule
LLaVA 1.5 7B & 18.5 & 4.2 & 5.1 & 0.2 \\
LLaVA 1.5 13B & 2.1 & 1.2 & 0.3 & 0.0 \\
Ovis2.5 9B & 22.1 & 3.8 & 4.0 & 0.0 \\
\bottomrule
\end{tabular}
\end{table}

\begin{table}[t]
\centering
\caption{\textbf{Top Geometric Confusion Pairs.} Target pairs sharing near-identical silhouette geometry that are trivially discriminable under HQ conditions. Models confuse each pair hundreds of times in silhouette mode despite high HQ accuracy, confirming the confusion is driven by geometric similarity rather than semantic ambiguity.}
\label{tab:confused_pairs}
\resizebox{\columnwidth}{!}{%
\small
\begin{tabular}{|l|c|r|}
\toprule
\rowcolor{shadecolor} \textbf{Confused Pair} & \textbf{HQ Acc (\%)} & \textbf{Sil Confusions} \\
\midrule
\multicolumn{3}{|c|}{\cellcolor{LGray2}\textbf{ImageNet-S} (\textit{n=100})} \\
\midrule
Admiral $\rightarrow$ Cabbage Butterfly & 71.0 & 380+ \\
Walker Hound $\rightarrow$ Beagle & 84.2 & 310+ \\
\midrule
\multicolumn{3}{|c|}{\cellcolor{LGray2}\textbf{WTP-Bench} (\textit{n=1,160})} \\
\midrule
Plusle $\rightarrow$ Minun & 82.5 & 540+ \\
Nidoran (Male) $\rightarrow$ Nidoran (Female) & 79.8 & 480+ \\
\midrule
\multicolumn{3}{|c|}{\cellcolor{LGray2}\textbf{CUB-200} (\textit{n=200})} \\
\midrule
Clark Nutcracker $\rightarrow$ American Crow & 68.4 & 210+ \\
Western Meadowlark $\rightarrow$ American Crow & 71.5 & 201+ \\
\midrule
\multicolumn{3}{|c|}{\cellcolor{LGray2}\textbf{DIS5K} (\textit{n=125})} \\
\midrule
Table+Chair $\rightarrow$ Chair & 68.2 & 172+ \\
Tricycle $\rightarrow$ Bicycle & 74.5 & 94+ \\
\midrule
\multicolumn{3}{|c|}{\cellcolor{LGray2}\textbf{PASCAL VOC} (\textit{n=20})} \\
\midrule
Bus $\rightarrow$ Car & 65.6 & 366+ \\
Train $\rightarrow$ Boat & 63.6 & 343+ \\
\midrule
\multicolumn{3}{|c|}{\cellcolor{LGray2}\textbf{ThinObject5K} (\textit{n=193})} \\
\midrule
Mop $\rightarrow$ Broom & 58.1 & 245+ \\
Comb $\rightarrow$ Hairbrush & 61.2 & 104+ \\
\bottomrule
\end{tabular}%
}
\end{table}

\noindent\textbf{Typological and Morphological Degradation.} Disaggregating failures by Pok\'emon elemental type reveals a clear geometric hierarchy of difficulty. Ghost-, Bug-, and Dragon-type silhouettes are consistently hardest: Ghost-types feature amorphous, non-contiguous floating geometry; Bug-types have multi-leg arrangements with thin appendages that lose discriminative structure at binary threshold; Dragon-types span a wide morphological range from serpentine to quadrupedal, collapsing into ambiguous blobs at low resolution. By contrast, quadrupedal Normal- and Ground-type forms preserve compact convex hulls with discriminative appendages (ears, tails), making them comparatively more identifiable. At the morphological level, Mega-evolved and Gigantamax forms are the hardest class in the entire benchmark: they add external geometric complexity on top of an already intricate base shape, simultaneously increasing silhouette ambiguity, compounding the baseline confusion between a form and its unevolved counterpart.

\begin{figure*}[t]
\centering
\begin{minipage}{0.48\textwidth}
    \centering
    \includegraphics[width=\linewidth]{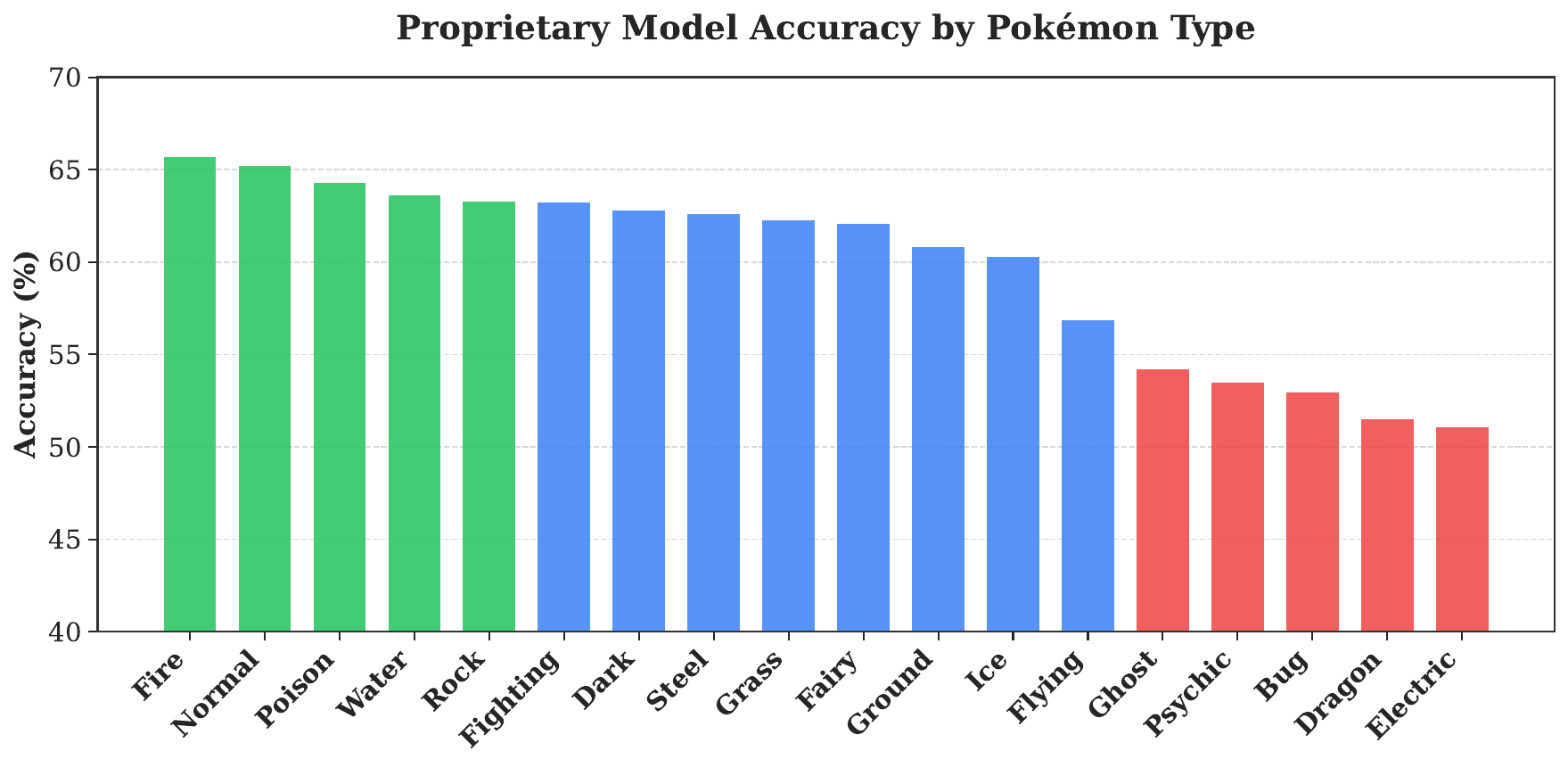}
\end{minipage}
\hfill 
\begin{minipage}{0.48\textwidth}
    \centering
    \includegraphics[width=\linewidth]{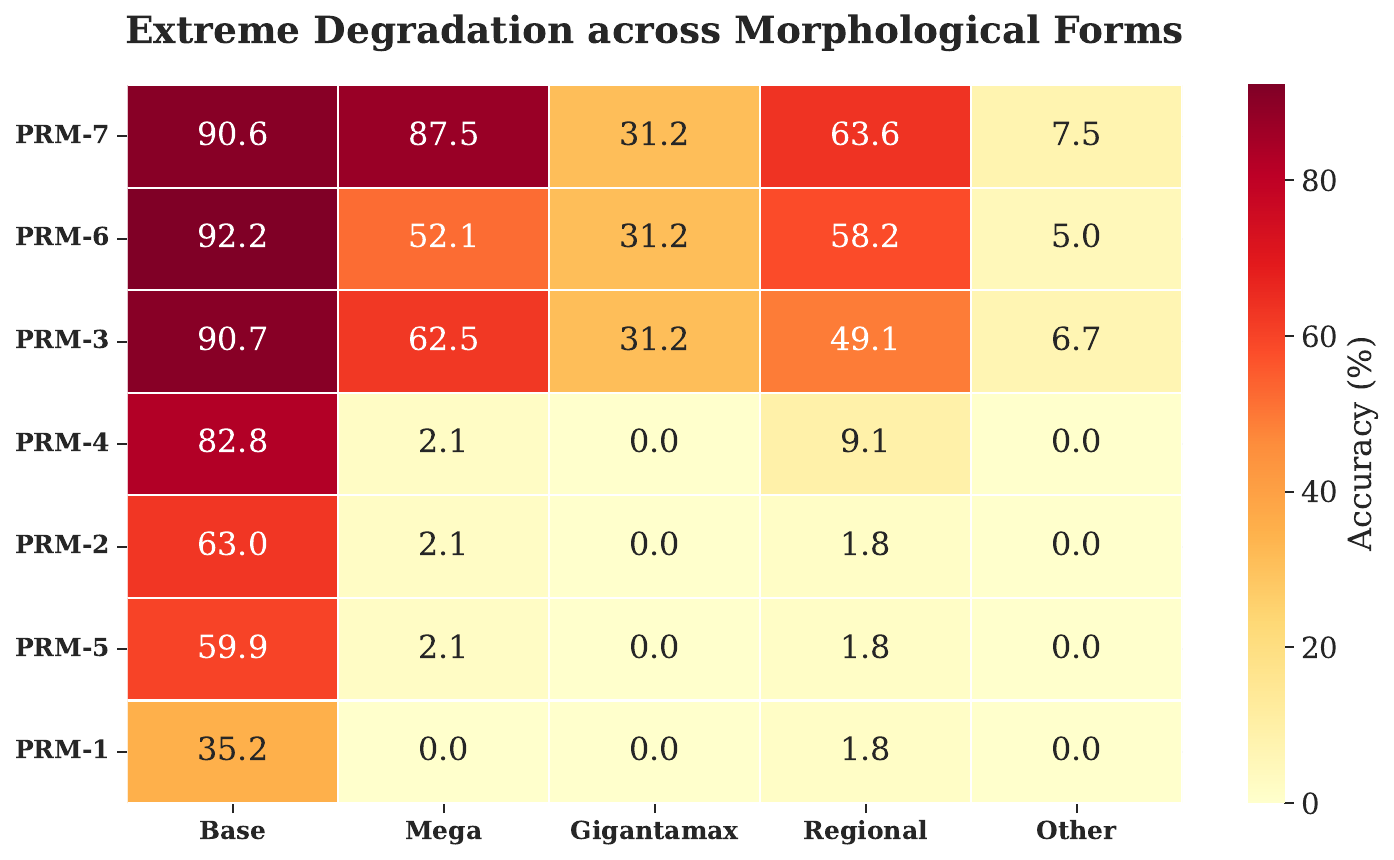}
\end{minipage}

\vspace{0.8em}
\caption{\textbf{Typological and Morphological Breakdown.} \textit{Left:} Accuracy stratified by Pokémon elemental type. Ghost-, Bug-, and Dragon-type silhouettes (amorphous geometry) are consistently hardest. \textit{Right:} Accuracy by morphological form. Alternate forms (Mega, Gigantamax) incur the steepest silhouette drops as added geometric complexity compounds the baseline difficulty.}
\label{fig:wtp_typology}
\end{figure*}

\noindent The elemental and morphological stratification in \Cref{fig:wtp_typology} confirms that texture removal amplifies pre-existing geometric ambiguity rather than introducing independent failures. Ghost- and Dragon-types carry the highest shape-based confusion rates because their irregular, non-convex geometries never had strong structural signatures to begin with; the Texture Bias Cliff merely exposes a deficit already latent in the encoder. By contrast, the steeper-than-average degradation on Mega and Gigantamax forms compared to their base counterparts directly implicates a deeper failure: the inability to represent geometric transformations. Mega evolution adds external appendages and enlarges specific body regions, yet models treat the Mega form as a wholly separate entity unrelated to its base, rather than as a structured geometric modification. The result is that Mega and Gigantamax forms inherit the confusion load of their base forms while adding new sources of silhouette ambiguity, compounding both the recognition difficulty and the hallucination rate. This observation directly motivates future work on relational geometric representations that encode inter-form structural correspondences rather than treating each silhouette as an independent flat template.

\section*{WTP-Bench Tiered Accuracy Analysis}
\label{sec:supp_tiers}

\Cref{tab:wtp_tiers} provides the complete breakdown of WTP-Bench accuracy across all three evaluation tiers for every model under both HQ and silhouette conditions. The Ultra Ball tier isolates form-variant confusion (predicting the correct species but wrong form), while the Great Ball tier captures evolutionary-family matches. Key findings: Claude Sonnet~4.5 shows the largest Ultra Ball gain on HQ ($65.09\% \to 79.22\%$, $+14.1$~pp), indicating strong species-level understanding but difficulty with alternate forms. Gemini 2.5 Pro achieves the highest Great Ball HQ accuracy ($91.38\%$), suggesting it has acquired a coarse evolutionary taxonomy from pre-training. On silhouettes, proprietary models gain ${\sim}12$--$15$~pp from Master to Great Ball, while open-weight models gain at most ${\sim}4$~pp---confirming near-total geometric blindness. SmolVLM shows \textit{zero} additional matches at any relaxed tier on HQ ($2.24\%$ at all three tiers), and Phi-3 Vision records $0.00\%$ across all silhouette tiers.

\begin{table}[t]
\centering
\caption{\textbf{WTP-Bench tiered accuracy (\%).} Performance under three progressively relaxed matching criteria---Master Ball (exact species + form), Ultra Ball (correct species, wrong form), and Great Ball (correct evolutionary lineage)---for both high-quality (HQ) and silhouette conditions. The Texture Bias Cliff persists across all tiers: even at the most lenient criterion, silhouette accuracy lags far behind HQ accuracy for every model.}
\label{tab:wtp_tiers}
\resizebox{\linewidth}{!}{%
\begin{tabular}{|c| rrr | rrr|}
\toprule
\rowcolor{shadecolor} & \multicolumn{3}{c|}{\textbf{High-Quality (Texture)}} & \multicolumn{3}{c|}{\textbf{Silhouette (Pure Shape)}} \\
\cmidrule(lr){3-5} \cmidrule(lr){5-7}
\rowcolor{shadecolor} \textbf{Model ID} & \textbf{Master} & \textbf{Ultra} & \textbf{Great} & \textbf{Master} & \textbf{Ultra} & \textbf{Great} \\
\midrule
\multicolumn{7}{|c|}{\cellcolor{LGray2}\textbf{Proprietary Models (PRM)}} \\
\midrule
PRM-1 & 27.59 & 35.78 & 42.07 & 27.41 & 35.86 & 41.90 \\
PRM-2 & 49.31 & 61.98 & 68.71 & 8.53  & 12.50 & 15.52 \\
PRM-3 & 77.24 & 83.02 & 86.21 & 30.00 & 35.95 & 42.07 \\
PRM-4 & 65.09 & 79.22 & 81.90 & 53.10 & 62.84 & 68.45 \\
PRM-5 & 46.90 & 59.05 & 63.71 & 7.93  & 11.38 & 14.14 \\
PRM-6 & 78.19 & 88.10 & 91.38 & 23.02 & 30.17 & 35.69 \\
PRM-7 & 78.97 & 84.22 & 84.66 & 49.48 & 58.53 & 62.50 \\
\midrule
\multicolumn{7}{|c|}{\cellcolor{LGray2}\textbf{Open-Weight Models $\mathbf{\leq 4B}$ (OSM 1--8)}} \\
\midrule
OSM-1  & 0.09  & 0.17  & 0.26  & 0.26  & 0.26  & 0.43  \\
OSM-2  & 2.24  & 4.57  & 6.03  & 0.78  & 2.67  & 3.28  \\
OSM-3  & 2.76  & 5.26  & 7.67  & 1.12  & 2.93  & 4.05  \\
OSM-4  & 2.24  & 2.24  & 2.24  & 0.95  & 0.95  & 1.03  \\
OSM-5  & 1.81  & 4.14  & 5.26  & 0.43  & 2.16  & 2.67  \\
OSM-6  & 2.84  & 5.00  & 6.55  & 0.60  & 2.41  & 3.10  \\
OSM-7  & 13.36 & 18.45 & 22.41 & 3.10  & 5.26  & 7.07  \\
OSM-8  & 0.17  & 1.03  & 1.64  & 0.00  & 0.00  & 0.00  \\
\midrule
\multicolumn{7}{|c|}{\cellcolor{LGray2}\textbf{Open-Weight Models $\mathbf{7B - 13B}$ (OSM 9--18)}} \\
\midrule
OSM-9  & 6.47  & 9.22  & 12.67 & 1.29  & 3.19  & 4.57  \\
OSM-10 & 3.53  & 5.78  & 7.16  & 0.78  & 2.33  & 3.10  \\
OSM-11 & 1.03  & 3.10  & 5.09  & 0.52  & 2.50  & 2.93  \\
OSM-12 & 3.53  & 6.38  & 9.91  & 1.81  & 3.97  & 5.69  \\
OSM-13 & 12.76 & 17.16 & 19.74 & 3.02  & 3.02  & 4.83  \\
OSM-14 & 2.93  & 5.86  & 8.45  & 0.78  & 2.50  & 3.10  \\
OSM-15 & 0.86  & 2.76  & 3.97  & 0.34  & 0.60  & 1.38  \\
OSM-16 & 5.95  & 9.31  & 13.88 & 2.16  & 4.57  & 6.72  \\
OSM-17 & 7.16  & 9.74  & 13.28 & 1.12  & 2.84  & 4.14  \\
OSM-18 & 1.81  & 4.74  & 6.90  & 0.69  & 2.41  & 2.76  \\
\midrule
\multicolumn{7}{|c|}{\cellcolor{LGray2}\textbf{Open-Weight Models $\mathbf{\geq 26B}$ (OSM 19)}} \\
\midrule
OSM-19 & 4.14  & 6.55  & 10.17 & 1.72  & 3.45  & 4.57  \\
\bottomrule
\end{tabular}%
}
\end{table}

\section*{Per-Dataset Analysis \& Extended Results}
\label{sec:supp_perdata}

\noindent\textbf{DIS5K.} On this geometrically intricate dataset, leading open-weight models suffer the largest absolute regressions: Qwen2-VL 7B drops from $78.77\%$ to $54.68\%$ on silhouettes, and LLaVA 1.5 13B regresses from $69.50\%$ to $39.34\%$---a consistent ${\sim}25$--$30$ percentage-point drop suggesting that ViT encoders struggle when faced with complex multi-contour boundaries without RGB cues.

\noindent\textbf{ThinObject5K.} This dataset presents a notable partial exception: mean performance drops from $47.5\% \pm 29.0\%$ to $46.5\% \pm 30.2\%$, an average drop of only $0.9\%$. Thin geometries such as wires, poles, and combs already present minimal interior texture, so silhouette conversion removes comparatively little discriminative signal. This exception is consistent with the Texture Bias Cliff hypothesis: the cliff only manifests strongly where textures previously carried discriminative weight.

\noindent\textbf{CUB-200.} Of 200 bird species, 137 achieve exactly $0\%$ silhouette accuracy across all 26 models. Beak curvature, tail length, and crest geometry alone are insufficient without plumage texture. This result directly validates the fine-grained inter-class challenge of CUB: silhouette-level bird anatomy is structurally ambiguous even to the highest-capacity models tested.

\noindent\textbf{WTP-Bench: Generation Bias Is Not Geometric.} The $79.2\%$ rate of Gen~1 hallucinations cannot be explained by geometric simplicity. Generation~1 and later generations exhibit statistically comparable silhouette complexity when controlling for form type (base, Mega, Gigantamax). Rather, the effect mirrors the human familiarity gradient: dedicated fans achieve $\sim$70\% on Gen~1 silhouettes but near single digits on later generations, driven by differential exposure rather than shape discriminability. This confirms the failure is rooted in pre-training corpus frequency, not objective geometric difficulty.

\Cref{tab:universally_hard} lists the most universally impenetrable targets: classes yielding exactly $0\%$ accuracy across \textit{all} 26 evaluated models. These entries share three recurring geometric archetypes: (1)~Gigantamax forms with extensive negative-space arms, tentacles, and floating disconnected geometry that defy continuous boundary tracking; (2)~regional variants (Alolan, Galarian) that differ from their base forms only in minor appendage repositioning invisible at silhouette scale; and (3)~near-radially-symmetric forms that collapse into indistinguishable blobs. The high HQ accuracies of these same targets (54--72\%) confirm that the classes are semantically recognizable under texture, ruling out label ambiguity as an explanation. \Cref{tab:wtp_generations} breaks WTP-Bench performance by generational tier. The consistent HQ--silhouette gap across all eight generations refutes the hypothesis that model failures stem from recent or obscure targets: even for Generation~1, where HQ accuracy exceeds 85\% for top proprietary models, silhouette accuracy collapses to under 60\%. The gradient from Generation~1 to Generation~8 in silhouette mode confirms that pre-training exposure, not geometric difficulty, is the dominant factor.

\begin{figure}[t]
\centering
\includegraphics[width=\linewidth]{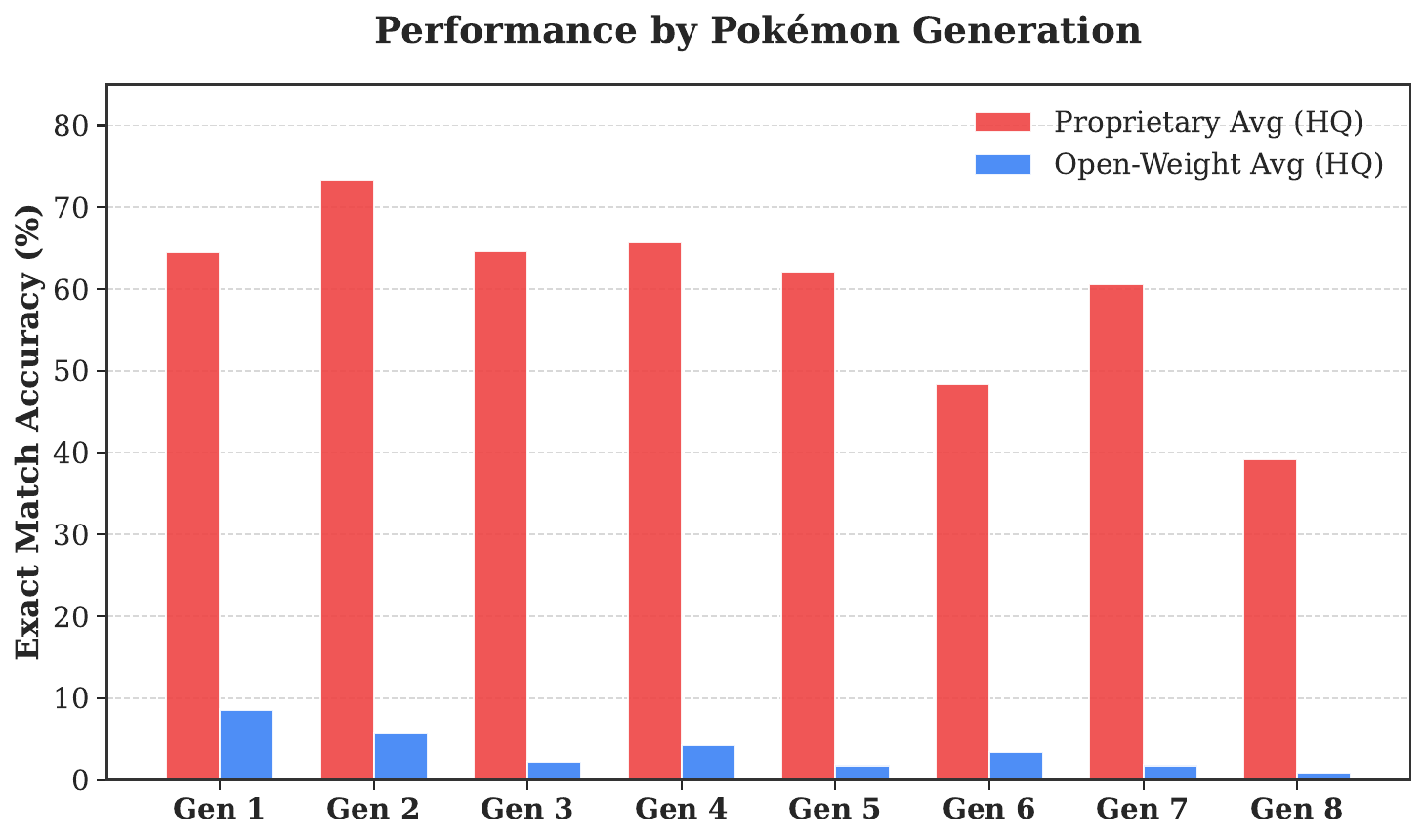}
\caption{\textbf{Pre-training Distribution Bias.} Both proprietary and open-weight models achieve their highest exact matches on Generation~1 targets, dropping sharply on modern targets with identical structural complexity.}
\label{fig:generation_bias}
\end{figure}

\begin{figure}[t]
\centering
\includegraphics[width=\linewidth]{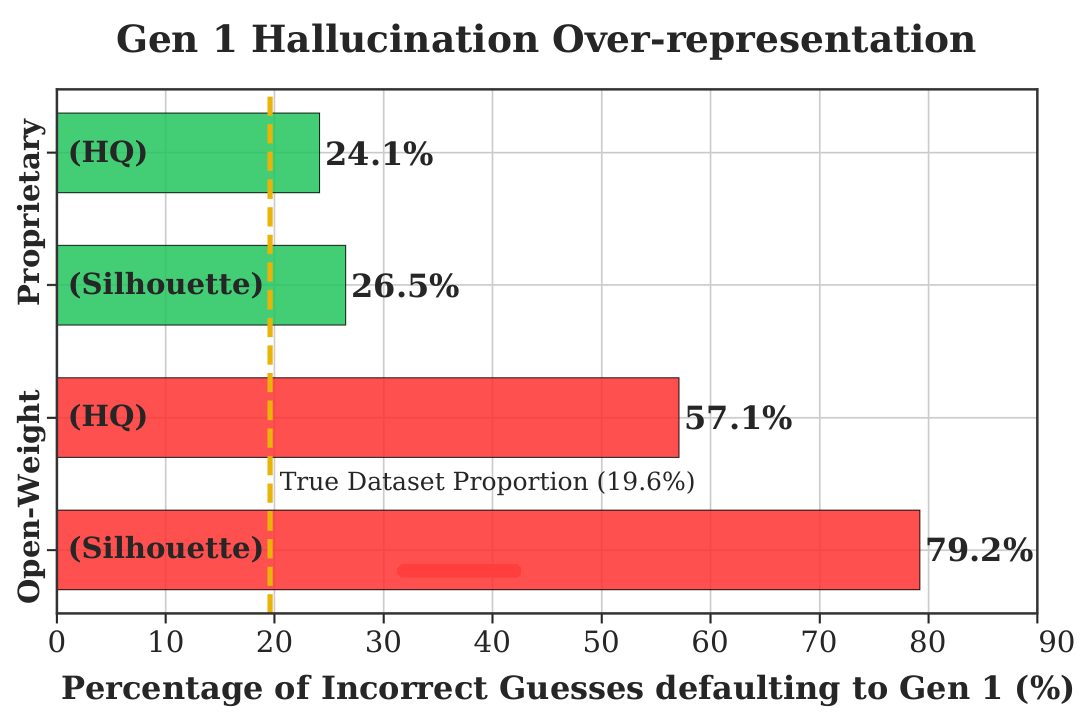}
\caption{\textbf{Generation~1 Over-representation in Failures.} When models fail on a silhouette, they hallucinate a Generation~1 target at a rate far exceeding the 19.6\% baseline prevalence.}
\label{fig:gen1_bias}
\end{figure}

\section*{Pre-Training Bias \& Hallucinations}
\label{sec:supp_hallucinations}

\begin{table}[t]
\centering
\small
\setlength{\tabcolsep}{4pt}
\begin{tabularx}{\linewidth}{|Y|rr|r|}
\toprule
\rowcolor{shadecolor} & \multicolumn{2}{c|}{\textbf{Open-Weight Silhouette Errors}} & \\
\cmidrule(lr){2-3}
\rowcolor{shadecolor}\textbf{Generation} & \textbf{Wrong Guesses} & \textbf{\% of Errors} & \textbf{Dataset \%} \\
\midrule
1 & 13,959 & \textbf{79.2\%} & 19.6\% \\
2 & 1,156 & 6.6\% & 9.6\% \\
3 & 613 & 3.5\% & 14.2\% \\
4 & 868 & 4.9\% & 11.2\% \\
5 & 371 & 2.1\% & 15.9\% \\
6 & 372 & 2.1\% & 9.4\% \\
7 & 174 & 1.0\% & 9.0\% \\
8 & 108 & 0.6\% & 11.1\% \\
\bottomrule
\end{tabularx}
\caption{\textbf{Hallucination Bias Distribution.} Frequency of incorrect predictions by the 16 open-weight models mapped to generational taxonomies, revealing a massive, statistically disproportionate bias towards Generation~1 textual priors when visual signals fail.}
\label{tab:gen1_bias}
\end{table}

When vision encoders are deprived of texture, the language decoder falls back on pre-training corpus statistics. \Cref{tab:gen1_bias} quantifies this behavior: hallucinated labels are overwhelmingly high-frequency, easily-recalled entities (\eg \textit{afghan hound}, \textit{american crow}, \textit{boat}) bearing little structural resemblance to the ground-truth silhouettes. \Cref{fig:generation_bias,fig:gen1_bias} extend this analysis to WTP-Bench, showing severe over-representation of Generation~1 targets in model failures despite Gen~1 comprising only 19.6\% of the evaluation set.

\section*{Extensive Error Analysis}
\label{sec:supp_errors}

\Cref{tab:confused_pairs} reveals the top geometric confusion pairs across all six datasets: instances where two structurally distinct targets are systematically conflated hundreds of times despite high HQ accuracy, confirming the confusion is driven by shared silhouette geometry rather than semantic ambiguity. These confusions are not random. Within WTP-Bench, top confusion pairs (Plusle/Minun, Nidoran Male/Female) share near-identical compact bilateral symmetry, differing only in minor ear orientation or tail length that collapses entirely at binary threshold resolution. Within ImageNet-S, confused butterfly species share globally symmetric forewings and rounded hindwings; within CUB-200, confused corvid species (Western Meadowlark, American Crow, etc) are indistinguishable at the 200~px silhouette scale. Across PASCAL VOC, the Bus/Car and Train/Boat confusions confirm that even basic category-level boundary geometry fails without canonical color cues.

\end{document}